\documentclass{article}
\usepackage{ijcai24}

\usepackage{times}
\usepackage{soul}
\usepackage{url}
\usepackage[hidelinks]{hyperref}
\usepackage[utf8]{inputenc}
\usepackage[small]{caption}
\usepackage{graphicx}
\usepackage{amsmath}
\usepackage{amsthm}
\usepackage{booktabs}
\usepackage{algorithm}
\usepackage{algorithmic}
\usepackage[switch]{lineno}
\usepackage{amssymb}
\usepackage{mathtools}
\usepackage{bm}
\usepackage{float}

\usepackage{booktabs}
\usepackage{stfloats}
\usepackage{longtable}
\usepackage{color}

\newcommand{\algo}{\textsf{\small MTRO}}
\newcommand{\derd}{\textsf{\small DERD}}
\newcommand{\barp}{\textsf{\small BARP}}

\title{Target Return Optimizer for Multi-Game Decision Transformer}

\author{
Kensuke Tatematsu$^1$\footnote{This work was conducted when the first author was interning at LY Corporation.}
\and
Akifumi Wachi$^2$ \and
\affiliations
$^1$Waseda University \space
$^2$LY Corporation\\
\emails
t.ken1358@akane.waseda.jp,
akifumi.wachi@lycorp.co.jp
}
\begin{document}

\maketitle
\begin{abstract}
Achieving autonomous agents with robust generalization capabilities across diverse games and tasks remains one of the ultimate goals in AI research.
Recent advancements in transformer-based offline reinforcement learning, exemplified by the Multi-Game Decision Transformer~\cite{MultigmaeDT}, have shown remarkable performance across various games or tasks.
However, these approaches depend heavily on human expertise, presenting substantial challenges for practical deployment, particularly in scenarios with limited prior game-specific knowledge.
In this paper, we propose an algorithm called Multi-Game Target Return Optimizer (\algo) to autonomously determine game-specific target returns within the Multi-Game Decision Transformer framework using solely offline datasets.
\algo~addresses the existing limitations by automating the target return configuration process, leveraging environmental reward information extracted from offline datasets.
Notably, \algo~does not require additional training, enabling seamless integration into existing Multi-Game Decision Transformer architectures. Our experimental evaluations on Atari games demonstrate that \algo~enhances the performance of RL policies across a wide array of games, underscoring its potential to advance the field of autonomous agent development.
\end{abstract}

\section{Introduction}
Reinforcement learning (RL) involves an agent interacting with an environment to learn a policy to maximize the expected cumulative reward.
RL is a powerful framework for solving sequential decision-making problems and has been applied to various applications such as alignment of large language models \cite{ouyang2022training}, recommendation systems \cite{afsar2022reinforcement,chen2019generative}, and robotics~\cite{plappert2018multi,wu2023learning}.
Offline RL \cite{offlineRL} is a paradigm for learning policies from a pre-collected dataset, which is particularly effective when environmental interaction is challenging due to cost or safety concerns. Despite promising developments in various approaches to offline RL, it is typically applied to solve a single task within the same environment.
Recently, with advances in foundational models~\cite{bommasani2021opportunities}, research in multi-game RL has progressed, aiming to apply a single policy model to various game environments characterized by different state-transition functions or reward functions.
This approach has been gaining interest as it seeks to acquire a general policy that can adapt to diverse environments from a variety of datasets.

Decision Transformer (DT, \cite{DT}) and its variants leverage the inference capabilities of transformer models ~\cite{transformer} by treating offline RL as a sequence modeling problem. This approach improves performance while addressing stability challenges associated with long-term credit assignments~\cite{sutton2018reinforcement}. Multi-Game DT extends the original DT to multi-game RL settings, enabling robust decision-making. Furthermore, similar to the scaling laws of large language models ~\cite{hoffmann2022training,kaplan2020scaling}, significant improvements have been observed over traditional offline RL methods such as conservative Q-learning (CQL, \cite{CQL}) and behavior cloning \cite{BehaviorCloaning}.

In both DT and Multi-Game DT, the learning process involves inputting the desired return into the transformer to learn a policy that generates conditionally actions. This framework allows control over the desired performance by adjusting the target return input during inference. However, determining the optimal target return to generate the best actions is challenging. In typical usage of the DT, the target return is manually set by humans. Additionally, Multi-Game DT addresses this by assuming a distribution over expert actions to generate optimal behaviors. While these methods are effective in a limited number of game environments, it becomes impractical to determine target returns based on human knowledge for multiple tasks in real-world scenarios.

This paper proposes an algorithm called Multi-Game Target Return Optimizer (\algo).
This algorithm automatically determines game-specific target returns from offline datasets, without relying on human knowledge. This method enhances the performance of pre-trained Multi-Game DT models across multiple games without additional training. Our contributions are summarized as follows:

\begin{itemize}
    \item We propose \algo~that consists of the following two components, which enable us to optimize game-specific target returns from offline datasets:
    \begin{enumerate}
        \item Data-driven Expert Return Distribution (\derd) that automatically determines distributions related to expert actions, replacing human assumptions.
        \item Bayes-Adjusted Return Prediction (\barp) that compares the predicted return distribution of the transformer with offline datasets and adjusts it based on predictive performance.
    \end{enumerate}
    \item We empirically demonstrate that \algo~improves performance in multi-game RL environments by obtaining game-specific target returns.
\end{itemize}

\section{Related Work}

Decision Transformer (DT,~\cite{DT}) is a promising approach that frames RL as a sequence modeling problem using the transformer architecture~\cite{transformer}.
DT demonstrates high performance due to the inference capabilities and scalability of the Transformer architecture, leading to extensive research in this area. Recent studies focus on advancements in the Transformer architecture and its representation methods~\cite{Starformer,Decisionconvformer,GeneralizedDT,janner2021offline}, as well as the integration of conventional RL techniques such as dynamic programming~\cite{Q-learningDT,Act:EmpoweringDT}. Beyond offline RL settings, research has explored fine-tuning DT online after pre-training offline~\cite{OnlineDT,offlinepre} and generalizing to unknown tasks from limited offline demonstration datasets~\cite{PromptingDT,HyperDT}. Additionally, there is growing interest in incorporating safety considerations into RL~\cite{ConstrainedDT}.

\paragraph{DT in multi-game RL.}
DT architectures have also been proven effective in multi-game RL settings.
Multi-Game DT~\cite{MultigmaeDT} extends the standard DT to multi-game RL settings, extracting knowledge from large volumes of offline datasets.
However, the diverse nature and complexity of games make creating a generalized policy challenging, necessitating the exploration of various approaches.
HarmoDT \cite{HarmoDT} identifies the optimal parameter space for each task, while Elastic DT \cite{ElasticDT} proposes a method to switch from suboptimal to optimal trajectories by adjusting the history length input to DT.

\paragraph{Target return in DT.}
In the DT paradigm, setting a target return as a condition for the transformer facilitates the generation of future actions that achieve the desired return.
The target return is an important element in determining the performance of actions and has been extensively studied.
Q-learning DT \cite{Q-learningDT} utilizes the outcomes of dynamic programming to relabel target returns in offline datasets for learning. Additionally, Multi-Game DT estimates target returns that generate expert actions by employing Bayes' rule during testing.
Return-Aligned DT \cite{Return-AlignedDT} not only aims to improve performance through target returns but also seeks to mitigate discrepancies between target returns and actual returns.

\section{Preliminaries}
We consider a sequential decision-making problem in which an agent interacts with multiple environments.
Each environment is modeled as a partially observable Markov decision process (POMDP), which is defined as a tuple
\begin{alignat}{2}
    \mathcal{M} \coloneqq \langle \mathcal{S}, \mathcal{A}, \mathcal{O}, \mathcal{P}, \mathcal{Z}, r \rangle,
\end{alignat}
where $\mathcal{S} \coloneqq \{s\}$ is a set of states, $\mathcal{A} \coloneqq \{a\}$ is a set of actions, $\mathcal{O} \coloneqq \{o\}$ is a set of observations, $\mathcal{P}$ is a state transition probability, $\mathcal{Z}$ is the observation density function meaning the probability of receiving observation $o_t \in \mathcal{O}$ from state $s_t \in \mathcal{S}$, and $r: \mathcal{S} \times \mathcal{A} \rightarrow \mathbb{R}$ is a reward function.
At each time step $t$, the agent in state $s_t \in \mathcal{S}$ receives an observation $o_{t} \sim \mathcal{Z}(\cdot \mid s_t)$ from the environment, selects an action $a_{t} \in \mathcal{A}$, and then receives a reward $r_{t} = r(s_t, a_t)$ and transits to a next state $s_{t+1} \sim \mathcal{P}(\cdot \mid s_t, a_t)$. 
Return of a trajectory $ {R}_{t} $ is defined as the cumulative sum of rewards from the time step $ t $ to the final step $ T $, expressed as $ {R}_{t} = \sum_{t'=t}^{T} r_{t'} $.

This paper aims to learn a single policy $\pi_\theta$ parameterized by $\theta$ that maximizes the future return ${R}_{t}$ in all environments where each environment is associated with each game.

\subsection{Decision Transformer (DT)}
DT~\cite{DT} is a paradigm that treats an RL problem as a sequence modeling problem, using the transformer architecture to derive effective policies from offline datasets.
Unlike traditional RL methods~\cite{DQN,A3C,PPO} that depend on value functions or policy gradients, this approach predicts future actions based on past sequence information.
In DT, a trajectory is defined as follows:
\begin{align}
    \tau = (R_{1},o_{1},a_{1},R_{2},o_{2},a_{2},...,R_{T},o_{T},a_{T}).
\end{align}
The DT model is trained on offline datasets to predict action $a_{t+1}$ based on a history of previous retuen $R_{\leq t}$, observation $o_{\leq t}$, and action $a_{\leq t}$.
During testing, the agent infers $a$ using observation $o$ obtained from the environment and the target return $\hat{R}$ specified by a human.

\subsection{Multi-Game Decision Transformer}

While standard DTs aim to solve a single game, the Multi-Game DT~\cite{MultigmaeDT}) explores multi-game settings, and aims to enable a single DT model to solve multiple games through learning from offline datasets. In Multi-Game DT, a sequence is defined as
\begin{align}
\tau = (o_{1}, {R}_{1}, a_{1}, r_{1}, o_{2}, {R}_{2}, a_{2}, r_{2}, \ldots, o_{T}, {R}_{T}, a_{T}, r_{T}).
\end{align}

Unlike the standard DT, Multi-Game DT utilizes information about the immediate reward $r$. The model's inputs are $o_{\leq t}$, $R_{\leq t}$, $a_{\leq t}$, and $r_{\leq t}$, while the outputs are $R_{t+1}$, $a_{t+1}$, and $r_{t+1}$. The prediction of future returns $R_{t+1}$ and rewards $r_{t+1}$ is conducted to acquire better representations. Additionally, the prediction of the return $R$ is used to set the target return $\hat{R}$ for outputting expert action sequences, as discussed in the next section. 

\subsection{Determining the Target Return}

The target return is vital for determining the quality of the generated action sequence. In DT, the interaction begins with a human specifying the desired performance as the initial target return $ \hat{R}_{1} $ for each game. The target return $ \hat{R}_{t} $ is iteratively updated at each step using the received reward $ r_{t} $ by 
\begin{equation}
\hat{R}_{t+1} = \hat{R}_{t} - r_{t}, \quad \forall{t \in [1, T-1]}.
\end{equation}
Empirical experiments in the original DT paper~\cite{DT} have shown a strong correlation between the target return and the actual observed episode return.

In DT, the training data was limited to those generated by medium-level or expert policies. However, in Multi-Game DT, non-expert data is utilized alongside expert data to learn a general-purpose agent.
Thus, Multi-Game DT utilizes Bayes’ rule with the return prediction distribution $P(R_{t} \mid \ldots)$ from the transformer to automatically generate the target return $\hat{R}$.
The target return $\hat{R}$ is sampled from the expert return distribution $ P(R_{t}, \ldots \mid \text{expert}_{t}) $ using the Bayes’ rule:
\begin{equation}
\label{eq:bayes}
P(R_{t}, \ldots \mid \text{expert}_{t}) \propto P(\text{expert}_{t} \mid R_{t}, \ldots) P(R_{t} \mid \ldots),
\end{equation}
where $ P(\text{expert}_{t} \mid R_{t}, \ldots) $ is a probability that the behavior is expert-level.
Existing studies, as represented by Multi-Game DT~\cite{MultigmaeDT}, typically assume a binary classifier using an inverse temperature $ \kappa$, with $ R_{\text{low}} $ and $ R_{\text{high}} $ as the lower and upper bounds of the return, which is defined as:

\begin{equation}
\label{eq:expert_probability}
P(\text{expert}_{t} \mid R_{t}, \ldots) \propto \exp\left(\kappa \left( \frac{R_t - R_{\text{low}}}{R_{\text{high}} - R_{\text{low}}} \right)\right).
\end{equation}
This assumption implies that the likelihood of a return sequence being classified as expert increases with larger target return $\hat{R}_t$. In other words, higher target returns are associated with more expert-like behavior sequences. This formulation enables the model to sample and set a target return $\hat{R}$ that is expected to lead to high performance.

\begin{algorithm*}[t]
\caption{Action Generation by \algo}
\label{algorithm}
\begin{algorithmic}[1]
\STATE $ P_{\mathrm{offline}}(\text{expert}_{t} \mid R_{t}, \ldots) \leftarrow \frac{n_{\mathrm{expert}}^{R_{t}}}{N(R_t)} $ 
\FOR{$ t = 1, 2, \ldots$ }
    \IF{$ t < \ell $}
        \STATE $\alpha(R_t) \leftarrow 1$
    \ELSIF{$t=\ell$}
        \STATE $\alpha(R_t) \leftarrow D_{\mathrm{KL}}\left(\frac{1}{\ell} \sum_{t'=1}^{\ell} P(R_{t'} \mid \ldots) \,\middle\vert\middle\vert\, \frac{N(\tilde{R}_1)}{N_\text{total}}\right) + 1$
    \ENDIF
    \STATE $\log P(R_{t}, \ldots \mid \text{expert}_{t}) \leftarrow \log P_{\mathrm{offline}}(\text{expert}_{t} \mid R_{t}, \ldots) + \frac{1}{\alpha(R_t)}\cdot \log P(R_{t} \mid \ldots) + \left(1-\frac{1}{\alpha(R_t)}\right) \cdot \log \frac {N(\tilde{R}_1)}{N_\text{total}}$
    \STATE Normalize $P(R_{t}, \ldots \mid \text{expert}_{t})$
    \STATE Sample a target return $ R_{t} \sim P(R_{t}, \ldots \mid \text{expert}_{t}) $
    \STATE Sample a next action $ a_{t} \sim P(a_{t} \mid R_{t}, \ldots) $
\ENDFOR
\end{algorithmic}
\end{algorithm*}

\paragraph{Challenges of the conventional method.}
However, this method has limitations when it comes to improving performance across different games. 
Applying Bayes' rule requires that the assumption $P(R_{t}, \ldots \mid \text{expert}_{t})$ matches the actual environment. Thus, using a single assumption across multiple environments with diverse reward structures, as in \eqref{eq:expert_probability}, poses challenges. Addressing these limitations could greatly enhance the effectiveness of multi-game RL policies across different environments.

\section{Method}
This paper proposes an algorithm called Multi-Game Target Return Optimizer (\algo). A key feature of \algo~is that the probability $ P(\text{expert}_t \mid R_t, \ldots) $ is derived directly from offline datasets, rather than relying on human assumptions as in \eqref{eq:expert_probability}.
This allows for improved performance across games without human knowledge, regardless of the reward structure of each game.
The pseudo-code of \algo~is outlined in Algorithm \ref{algorithm}.

In this paper, we apply \algo~under an assumption that a pre-trained Multi-Game DT model is available a priori.
The main objective of \algo~is to improve performance in multiple games without additional training by determining more appropriate target returns for Multi-Game DT models.

\begin{figure*}[t]
    \centering
    \includegraphics[width=\textwidth]{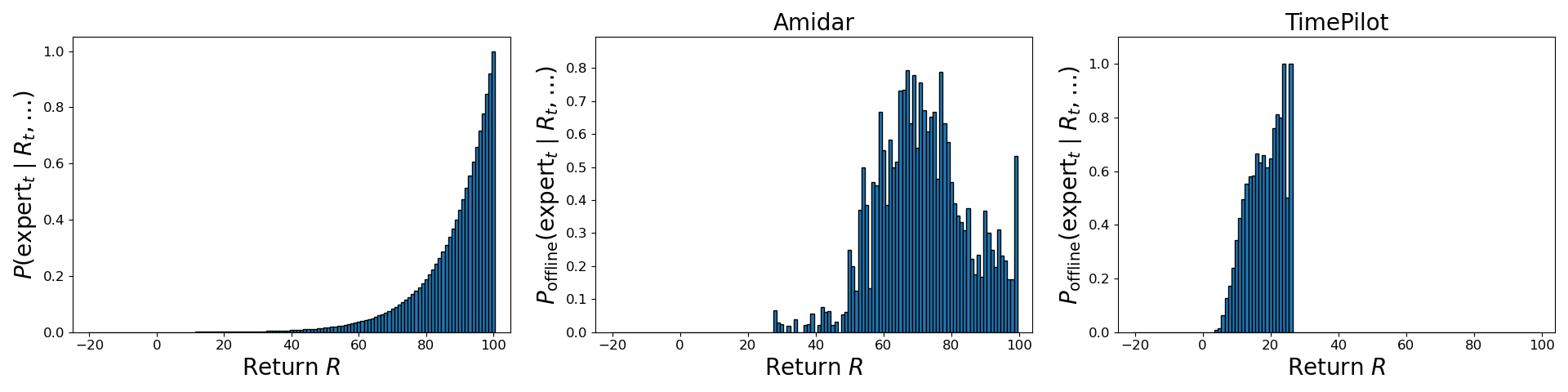}
    \caption{This figure compares distributions $P(\text{expert}_{t} \mid R_{t} \ldots)$ derived from \eqref{eq:expert_probability} and \eqref{eq:expert_offline}. \textbf{Left}: $P(\text{expert}_{t} \mid R_{t}, \ldots)$ under the assumption of \eqref{eq:expert_probability}, which increases exponentially from $R_{\text{low}}$ to $R_{\text{high}}$ for the quantized returns in the range [-20, 100]. \textbf{Middle}: $P_\text{offline}(\text{expert}_{t} \mid R_{t}, \ldots)$ based on \eqref{eq:expert_offline} for the Amidar game, where the distribution is closer to a normal distribution rather than an exponential type. \textbf{Right}: $P_\text{offline}(\text{expert}_{t} \mid R_{t}, \ldots)$ based on \eqref{eq:expert_offline} for the TimePilot game, showing a distribution closer to an exponential type, but the maximum return value is significantly smaller than that in the left figure.}
    \label{fig:p(expert|R)}
\end{figure*}

\subsection{DERD: Data-driven Expert Return Distribution}
\label{chap:DARD}

Traditionally, Multi-Game DT has applied the standard Bayes’ rule ~\eqref{eq:bayes} based on the assumption represented as~\eqref{eq:expert_probability}.
This equation suggests that the probability of being an expert increases as the return approaches $ R_\text{high} $. However, $ R_\text{high} $ defines the maximum episode return across all games, but not all games have the episode in the offline datasets that reaches $ R_\text{high} $. Furthermore, due to the various reward structures present in different games, the distribution of $ P(\text{expert}_t \mid R_t, \ldots) $ may not naturally conform to an exponential form.

To address these challenges, we calculate $ P_\text{offline}(\text{expert}_t \mid R_t, \ldots) $ using episodic return information from offline datasets.
In this approach, the target return can be set individually for each game, reflecting the unique reward structures of each game; thus, we can enhance the probability of generating expert-level state action sequences within the context of specific games.
Specifically, each return is quantized to an integer value $\tilde{R}$, allowing us to approximately compute $ P(\text{expert}_t \mid R_t, \ldots) $ as follows:
\begin{equation}
\label{eq:expert_offline}
P_{\mathrm{offline}}(\text{expert}_t \mid R_t, \ldots) \propto \frac{n_{\mathrm{expert}}^{\tilde{R}_t}}{N(\tilde{R}_t)},
\end{equation}
where $ n_{\mathrm{expert}}^{\tilde{R}_t} $ represents the number of episodes with quantized return $ \tilde{R}_t $ labeled as expert data, and $ N(\tilde{R}_t)$ represents the number of episodes with return $ \tilde{R}_t $.

This formula represents the probability that a quantized return $ \tilde{R}_t $ is classified as expert within the offline datasets. By calculating it for each game, this approach aims to capture the different game-specific reward structures, enhancing the accuracy of expert classification. This method is simple and easy to implement, as it requires calculating the probability only once before testing, and then simply replacing it with \eqref{eq:expert_probability}.

Figure~\ref{fig:p(expert|R)} compares the results of calculations from \eqref{eq:expert_probability} and \eqref{eq:expert_offline}. Figure~\ref{fig:p(expert|R)} (left) shows the values computed using \eqref{eq:expert_probability}, which is used in all games in the Multi-Game DT. On the other hand, Figure~\ref{fig:p(expert|R)} (Middle, Right) presents the results derived from \eqref{eq:expert_offline} based on offline datasets, highlighting the differences from the exponential type. Our method allows for the use of $ P_{\mathrm{offline}}(\text{expert}_t \mid R_t, \ldots) $ that considers the difficulty and reward structure of each game.

\subsection{BARP: Bayes-Adjusted Return Prediction}
\label{chap:BARP}

\begin{figure*}[t]
    \centering
    \includegraphics[width=\textwidth]{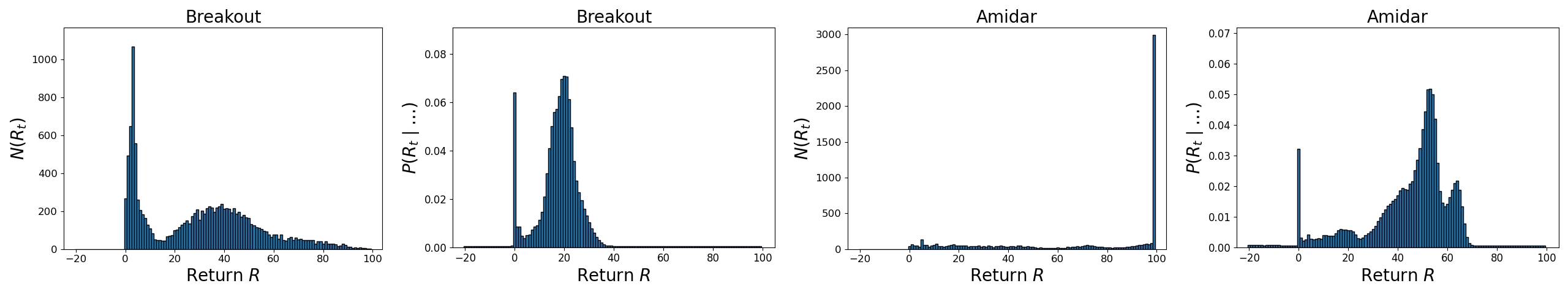}
\caption{This figure illustrates the differences between $N(R_1)$ derived from offline datasets, and the transformer's output $P(R_{t} \mid \ldots)$. \textbf{Left}: In the Breakout Game, $P(R_{t} \mid \ldots)$ tends to lean towards lower $R$ values compared to $N(R_t)$, suggesting that lower returns are more probable than what the offline datasets indicates. \textbf{Right}: In the Amidar Game, while $N(R_t)$ peaks at higher $R$ values, $P(R_{t} \mid \ldots)$ shows a notably different pattern, highlighting a divergence from the expected distribution.}
    \label{fig:p(R|...)}
\end{figure*}

In Multi-Game DT, which learns multiple games with different reward designs using a single transformer model, it is also challenging to accurately predict the distribution of returns $ P(R_{t} \mid \ldots) $ from the model's output. When predictions of $ P(R_{t} \mid \ldots) $ are inaccurate, we cannot generate valid target returns in \eqref{eq:bayes} due to inaccurate $P(R_{t}, \ldots \mid \text{expert}_{t})$. Therefore, we modify \eqref{eq:bayes} while incorporating the prediction accuracy of returns as follows: 
\begin{align}
\label{eq:proposed}
\log P(R_{t}, \ldots \mid \text{expert}_{t})
&\propto  \log P(\text{expert}_{t} \mid R_{t}, \ldots) \nonumber \\
&\quad + \frac{1}{\alpha(R_t)}\cdot \log P(R_{t} \mid \ldots) \\
&\quad + \left(1-\frac{1}{\alpha(R_t)}\right) \cdot \log \frac {N(\tilde{R}_t)}{N_\text{total}}, \nonumber
\end{align}
where $N_\text{total}$ represents the total number of episodes in the offline datasets, and $N(\tilde{R}_t)/{N_\text{total}}$ represents the probability distribution over the offline datasets for each $\tilde{R}_t$.
The function $\alpha(R_t)$ is defined based on the KL-divergence between the average of the predicted distribution $P(R_{t=0 \sim \ell} \mid \ldots)$ from the first few $\ell$ frames during inference and $N(\tilde{R}_t)/{N_\text{total}}$ with respect to $R_t$.
While we use $\alpha(R_t) = 1$ for all $t < \ell$, we define $\alpha(R_t)$ for all $t \ge \ell$ as follows:
\begin{align}
\label{eq:alpha}
    \alpha(R_t) \coloneqq D_{\mathrm{KL}}\left(\frac{1}{\ell} \sum_{t=1}^{\ell} P(R_{t} \mid \ldots) \,\middle\vert\middle\vert\, \frac{N(\tilde{R}_1)}{N_\text{total}}\right) + 1.
\end{align}

When $\alpha(R_t) = 1$ (i.e., the KL-divergence is equal to $0$), the predicted return distribution $P(R_{t} \mid \ldots)$ is considered accurate, and the standard Bayes' rule~\eqref{eq:bayes} is applied. If the KL-divergence is greater than $0$, $P(R_{t} \mid \ldots)$ is adjusted by adding the distribution of the target return from the offline datasets, ${N(\tilde{R}_t)}/{N_\text{total}}$, in \eqref{eq:proposed}.

Figure~\ref{fig:p(R|...)} compares $N(R_t)$ calculated from offline datasets with $P(R_{t} \mid \ldots)$ obtained during the inference of a Multi-Game DT model. The significant discrepancy between these distributions indicates poor predictive accuracy of $P(R_{t} \mid \ldots)$, suggesting that Bayes' theorem does not hold. We address this issue by using \eqref{eq:proposed}.

\section{Experiment}

In this paper, we conduct empirical evaluations to assess the performance of our \algo.
First, we evaluate its performance across various games in comparison with other baselines.
Next, we show that the two components of \algo~(i.e., \derd~and \barp) lead to more reasonable target return distributions $P(R_t \mid \text{expert}_t)$ than simply using the Bayes' rule~\eqref{eq:bayes}.

\subsection{Problem Settings}

We evaluate \algo~using the Atari benchmark~\cite{ATARI,atari2} as with the Multi-Game DT. In this paper, we use the DQN Replay Dataset~\cite{dqnreplaydataset} as offline datasets. This comprehensive dataset comprises Atari game trajectories where rewards are clipped to $[-1, 1]$ and the returns are quantized to the range $\{-20, 100\}$. It includes 50 million steps of experience for each of the five DQN agents~\cite{DQN} recorded during their learning processes.
In our experiments, we focus on 39 of the 41 games learned by Multi-Game DT, for which the score achieved by a human player is publicly available. 

\subsection{Metrics}
The performance of individual Atari games is measured using the Human Normalized Score (HNS,~\cite{HumanlevelCT}), calculated with the following formula:

\begin{equation}
\text{HNS} = \frac{\text{score} - \text{score}_{\text{random}}}{\text{score}_{\text{human}} - \text{score}_{\text{random}}},
\end{equation}
where $\text{score}_{\text{random}}$ is the score obtained by playing the game with random actions and $\text{score}_{\text{human}}$ is the score achieved by a human player~\cite{humanrandomscore}. To create an aggregate comparison metric across all games, we use the Interquartile Mean (IQM,~\cite{Agarwal2021DeepRL}) of the human normalized scores across all games. The IQM is computed after conducting 50 trials for each game, with a single model.

\subsection{Implementation of \algo}

We implemented \algo~based on the publicly available pre-trained model of Multi-Game DT\footnote{\url{https://sites.google.com/view/multi-game-transformers}}, which has been trained on 41 Atari games and consists of 200 million parameters. By adding two key techniques of \derd~and \barp, we improved performance without additional training. 

Specifically, for implementing \derd, we calculated $N(R_t)$ and $n_{\mathrm{expert}}^{R_t}$ from offline datasets. This involved utilizing experience from a single DQN agent within the DQN replay dataset, as used in Multi-Game DT.  We categorized the data for each game, designating the final 10\% episodes of the experience as expert level.
Additionally, we computed the KL-divergence as in \eqref{eq:alpha}. For this, we saved $P(R_{t} \mid \ldots)$ from the first 20 steps during inference (i.e., $\ell = 20$). Starting from the 21st step, we applied \eqref{eq:proposed} to determine the target return.

\subsection{Baseline Methods}

We also implemented four baseline methods.

\paragraph{Multi-Game DT.}

Multi-Game DT utilizes 100 million steps of experience from two agents, derived from the DQN Replay Dataset, for each of the 41 games. This dataset encompasses actions from all stages of learning, containing both expert and non-expert data. Consequently, when aiming to generate expert-like behavior, Multi-Game DT applies Bayes' theorem~\eqref{eq:bayes}. 

\paragraph{Na\"ive method.} 
As another baseline method, we utilize a straightforward approach using 
\begin{equation}
\begin{aligned}
\label{propose2}
\log P(R_{t}, \ldots \mid \text{expert}_{t}) &\propto \log P_{\mathrm{offline}}(\text{expert}_{t} \mid R_{t}, \ldots) \\
&\quad + \frac{1}{\alpha(R_t)} \cdot \log P(R_{t} \mid \ldots).
\end{aligned}
\end{equation}
This method serves as an alternative to correcting $ P(R_{t} \mid \ldots) $ with \eqref{eq:proposed}. The function $\alpha(R_t)$ is defined as with \eqref{eq:proposed}. When $\alpha (R_t)$ is 1 (i.e., the KL-divergence is equal to $0$), the standard Bayes' rule is applied. If $\alpha (R_t)$ is greater than 1, we address the prediction accuracy issue of $ P(R_{t} \mid \ldots) $ by increasing the influence of $P_{\mathrm{offline}}(\text{expert}_t \mid R_t, \ldots)$.
We can also interpret that \eqref{propose2} is a simplified version of \algo~where the last term of \eqref{eq:proposed} is ignored.

\paragraph{Two baselines for ablation studies.}

We also implement two baseline methods where either \derd~and~\barp~is removed from \algo~in order to show that their combinations are needed to enhance the performance of \algo.

\subsection{Main Results}

We compare the performance of \algo~and four baseline methods to investigate the effectiveness of \algo.
Table~\ref{fig:HNSscore} shows the average HNS for each method, along with the mean, the standard deviation, and the overall IQM.
\algo, using both \derd~ and \barp, achieves the highest IQM score with 28\% improvement over Multi-Game DT. Notably, the Na\"ive method and \algo~ (w/o \barp), both using \derd, also demonstrate strong performance, confirming the effectiveness of \derd.
A table summarizing raw scores is presented in Table~\ref{tab:raw_score} as a supplementary material.

To further evaluate the effectiveness of \algo, we compared the performance of each method against the Multi-Game DT on a per-game basis. We focused on identifying games where performance improved or declined significantly. Table~\ref{fig:performance_comparison} shows that \algo~consistently achieves high scores

\clearpage
\begin{table*}[t]
\caption{Results for the Atari games, including the Human Normalized Score (HNS) with mean, standard deviation, and overall IQM over 50 trials. Bold text indicates the highest score for each game and the highest IQM score for the algorithm.}
\begin{center}
\begin{tabular}{lrrrrr}
\toprule
Game Name& Multi-Game DT & Na\"ive method & \algo~(w/o \barp) & \algo~(w/o \derd) & \algo \\
\midrule
Amidar & 0.081 $\pm$ 0.046 & 0.093 $\pm$ 0.047 & 0.095 $\pm$ 0.051 & \textbf{0.097 $\pm$ 0.045} & 0.080 $\pm$ 0.030 \\
Assault & 3.626 $\pm$ 1.647 & 3.901 $\pm$ 1.749 & 3.732 $\pm$ 1.788 & \textbf{4.003 $\pm$ 1.745} & 3.463 $\pm$ 1.768 \\
Asterix & 0.831 $\pm$ 0.232 & 0.998 $\pm$ 0.312 & \textbf{1.019 $\pm$ 0.486} & 0.871 $\pm$ 0.250 & 0.984 $\pm$ 0.376 \\
Atlantis & 3.935 $\pm$ 9.934 & \textbf{15.547 $\pm$ 19.076} & \textbf{15.547 $\pm$ 19.076} & -0.274 $\pm$ 0.201 & 14.769 $\pm$ 19.657 \\
BankHeist & -0.009 $\pm$ 0.013 & -0.010 $\pm$ 0.012 & \textbf{-0.001 $\pm$ 0.021} & -0.008 $\pm$ 0.018 & -0.010 $\pm$ 0.012 \\
BattleZone & 0.346 $\pm$ 0.194 & 0.425 $\pm$ 0.170 & \textbf{0.451 $\pm$ 0.209} & 0.368 $\pm$ 0.191 & 0.425 $\pm$ 0.178 \\
BeamRider & 0.461 $\pm$ 0.286 & 0.462 $\pm$ 0.257 & 0.463 $\pm$ 0.252 & \textbf{0.468 $\pm$ 0.244} & 0.451 $\pm$ 0.268 \\
Boxing & 7.600 $\pm$ 0.484 & 7.595 $\pm$ 0.478 & 7.580 $\pm$ 0.538 & \textbf{7.677 $\pm$ 0.427} & 7.675 $\pm$ 0.418 \\
Breakout & 8.494 $\pm$ 2.864 & 10.363 $\pm$ 2.894 & 10.592 $\pm$ 3.360 & 8.387 $\pm$ 3.767 & \textbf{11.306 $\pm$ 2.756} \\
Centipede & 0.023 $\pm$ 0.167 & 0.064 $\pm$ 0.158 & \textbf{0.082 $\pm$ 0.193} & 0.077 $\pm$ 0.186 & 0.054 $\pm$ 0.148 \\
ChopperCommand & \textbf{0.147 $\pm$ 0.139} & 0.101 $\pm$ 0.136 & 0.102 $\pm$ 0.136 & 0.048 $\pm$ 0.115 & 0.087 $\pm$ 0.132 \\
CrazyClimber & 4.408 $\pm$ 0.762 & 4.618 $\pm$ 0.772 & 4.618 $\pm$ 0.772 & \textbf{4.698 $\pm$ 0.711} & 4.618 $\pm$ 0.772 \\
DemonAttack & \textbf{6.920 $\pm$ 4.035} & 5.764 $\pm$ 3.116 & 5.910 $\pm$ 3.464 & 6.175 $\pm$ 3.687 & 6.390 $\pm$ 5.460 \\
DoubleDunk & 2.182 $\pm$ 2.161 & 1.564 $\pm$ 2.329 & \textbf{2.545 $\pm$ 2.294} & 1.027 $\pm$ 1.960 & 1.345 $\pm$ 2.349 \\
Enduro & 1.502 $\pm$ 0.321 & 1.620 $\pm$ 0.373 & 1.620 $\pm$ 0.373 & \textbf{1.661 $\pm$ 0.313} & 1.620 $\pm$ 0.373 \\
FishingDerby & \textbf{1.689 $\pm$ 0.648} & 1.603 $\pm$ 0.697 & 1.475 $\pm$ 0.738 & 1.573 $\pm$ 0.734 & 1.517 $\pm$ 0.739 \\
Freeway & 1.009 $\pm$ 0.057 & 1.003 $\pm$ 0.053 & 1.005 $\pm$ 0.053 & 0.845 $\pm$ 0.065 & \textbf{1.010 $\pm$ 0.052} \\
Frostbite & \textbf{0.497 $\pm$ 0.201} & 0.129 $\pm$ 0.145 & 0.108 $\pm$ 0.134 & 0.341 $\pm$ 0.199 & 0.190 $\pm$ 0.149 \\
Gopher & 4.460 $\pm$ 2.397 & 5.321 $\pm$ 3.027 & 4.229 $\pm$ 2.591 & 5.309 $\pm$ 3.166 & \textbf{5.550 $\pm$ 3.822} \\
Gravitar & \textbf{-0.035 $\pm$ 0.040} & -0.041 $\pm$ 0.034 & -0.040 $\pm$ 0.034 & -0.041 $\pm$ 0.029 & -0.042 $\pm$ 0.033 \\
Hero & 0.608 $\pm$ 0.085 & 0.629 $\pm$ 0.078 & 0.643 $\pm$ 0.044 & \textbf{0.648 $\pm$ 0.031} & 0.647 $\pm$ 0.032 \\
IceHockey & 0.152 $\pm$ 0.275 & 0.274 $\pm$ 0.289 & \textbf{0.286 $\pm$ 0.288} & 0.043 $\pm$ 0.331 & 0.218 $\pm$ 0.319 \\
Jamesbond & 1.713 $\pm$ 0.981 & \textbf{2.389 $\pm$ 0.812} & 2.122 $\pm$ 0.743 & 1.315 $\pm$ 1.049 & 2.235 $\pm$ 0.742 \\
Kangaroo & \textbf{3.933 $\pm$ 0.963} & 3.871 $\pm$ 1.107 & 3.812 $\pm$ 1.181 & 1.699 $\pm$ 1.314 & 3.763 $\pm$ 0.989 \\
Krull & 1.019 $\pm$ 0.733 & -0.288 $\pm$ 0.462 & 0.985 $\pm$ 0.931 & \textbf{6.378 $\pm$ 1.225} & 6.348 $\pm$ 1.303 \\
KungFuMaster & 0.662 $\pm$ 0.287 & \textbf{0.706 $\pm$ 0.269} & 0.618 $\pm$ 0.227 & 0.663 $\pm$ 0.259 & 0.642 $\pm$ 0.220 \\
NameThisGame & 0.940 $\pm$ 0.269 & 0.943 $\pm$ 0.292 & 0.943 $\pm$ 0.292 & \textbf{1.032 $\pm$ 0.272} & 0.943 $\pm$ 0.292 \\
Phoenix & 0.649 $\pm$ 0.075 & 0.655 $\pm$ 0.065 & 0.643 $\pm$ 0.135 & \textbf{0.663 $\pm$ 0.076} & 0.638 $\pm$ 0.074 \\
Qbert & 0.444 $\pm$ 0.239 & 0.381 $\pm$ 0.200 & 0.339 $\pm$ 0.222 & \textbf{0.467 $\pm$ 0.229} & 0.458 $\pm$ 0.251 \\
Riverraid & 0.846 $\pm$ 0.219 & 0.817 $\pm$ 0.233 & 0.843 $\pm$ 0.209 & 0.864 $\pm$ 0.218 & \textbf{0.893 $\pm$ 0.219} \\
RoadRunner & 3.607 $\pm$ 1.060 & 3.360 $\pm$ 1.011 & 3.106 $\pm$ 1.144 & 3.221 $\pm$ 0.802 & \textbf{3.775 $\pm$ 1.237} \\
Robotank & 4.860 $\pm$ 1.034 & 4.854 $\pm$ 0.998 & \textbf{4.934 $\pm$ 0.901} & 4.324 $\pm$ 1.433 & 4.695 $\pm$ 1.220 \\
Seaquest & \textbf{0.110 $\pm$ 0.037} & 0.094 $\pm$ 0.051 & 0.069 $\pm$ 0.043 & 0.052 $\pm$ 0.043 & 0.090 $\pm$ 0.041 \\
TimePilot & -0.632 $\pm$ 1.032 & \textbf{-0.195 $\pm$ 0.986} & -0.502 $\pm$ 0.978 & -0.555 $\pm$ 0.991 & -0.448 $\pm$ 0.984 \\
UpNDown & 1.230 $\pm$ 0.564 & 1.172 $\pm$ 0.483 & 1.013 $\pm$ 0.489 & 1.156 $\pm$ 0.564 & \textbf{1.256 $\pm$ 0.562} \\
VideoPinball & -11.419 $\pm$ 0.283 & -11.324 $\pm$ 0.391 & -11.374 $\pm$ 0.340 & \textbf{-11.253 $\pm$ 0.570} & -11.296 $\pm$ 0.509 \\
WizardOfWor & -0.049 $\pm$ 0.091 & \textbf{-0.035 $\pm$ 0.100} & -0.050 $\pm$ 0.067 & -0.044 $\pm$ 0.087 & -0.042 $\pm$ 0.081 \\
YarsRevenge & 0.243 $\pm$ 0.169 & 0.246 $\pm$ 0.172 & 0.225 $\pm$ 0.175 & 0.236 $\pm$ 0.155 & \textbf{0.283 $\pm$ 0.207} \\
Zaxxon & 0.022 $\pm$ 0.046 & 0.014 $\pm$ 0.027 & 0.005 $\pm$ 0.016 & 0.003 $\pm$ 0.017 & \textbf{0.030 $\pm$ 0.038} \\
\midrule
IQM &1.35 & 1.59 & 1.61 &  1.37 & \textbf{1.73} \\
\bottomrule
\label{fig:HNSscore}
\end{tabular}
\end{center}
\end{table*}

\clearpage

\noindent
across various games, demonstrating fewer declines and more improvements compared to other methods.

\begin{figure*}[t]
    \centering
    \includegraphics[width=\textwidth]{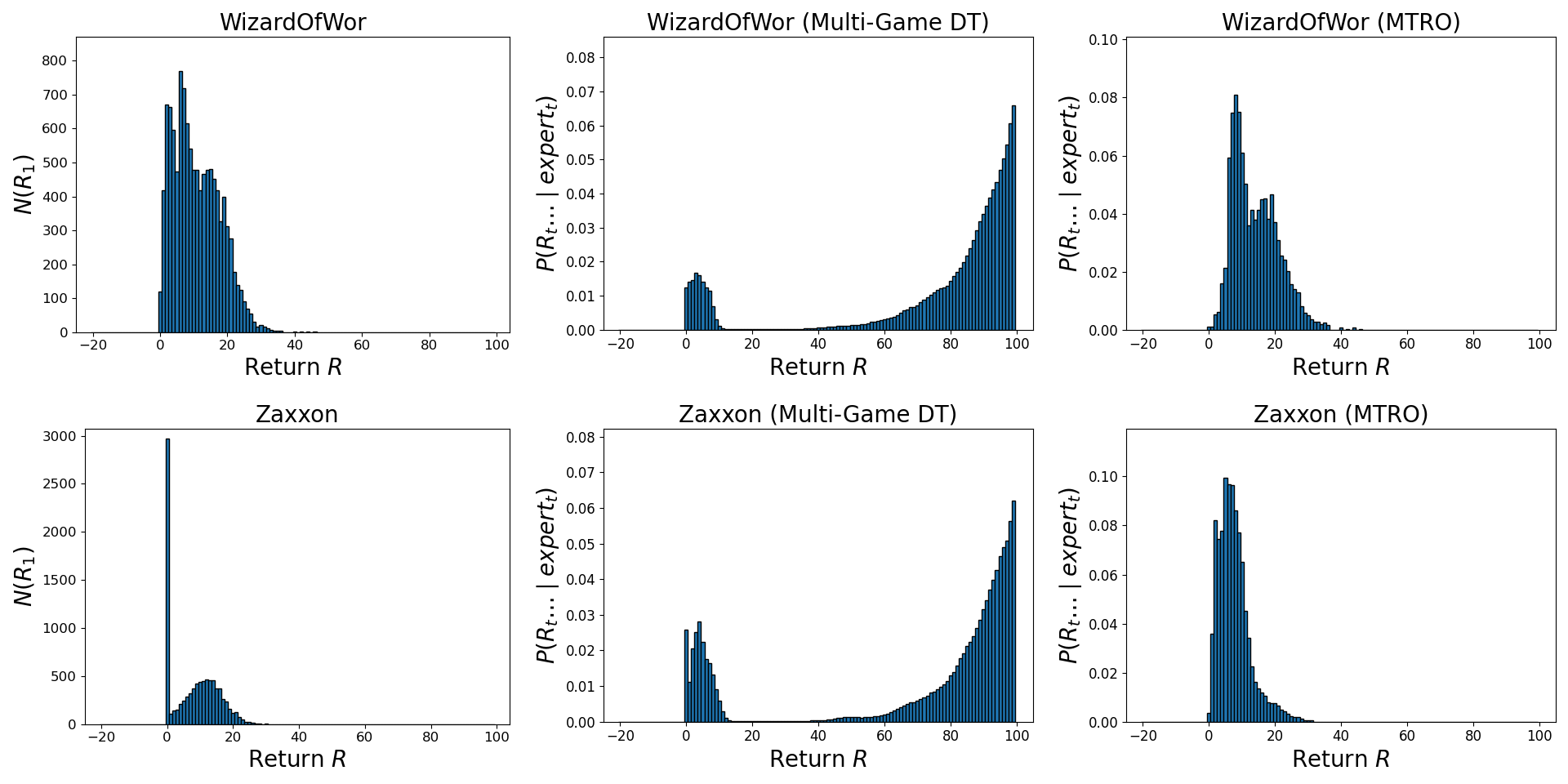}
\caption{This figure demonstrates the effectiveness of target returns in \algo~by comparing the frequency of episode returns from offline datasets, $ N(R_1) $, with $P(R_{t}, \ldots \mid \text{expert}_{t})$ used for sampling target returns in both Multi-Game DT and \algo. Here, $P(R_{t}, \ldots \mid \text{expert}_{t})$ represents the distribution at the beginning of episodes in the games. In both games, it is evident that Multi-Game DT might select target returns in the range $ R = 60 \sim 100 $, which are not included in $ N(R_1) $. On the other hand, because \algo~samples feasible target returns based on offline datasets, it is suitable as input to the transformer and is expected to generate expert actions.}
    \label{fig:exp_target_return}
\end{figure*}

The reason for the improved performance by \algo~ is to set feasible target returns based on offline datasets, unlike Multi-Game DT which utilizes distributions significantly different from the features of offline datasets.
In Figure~\ref{fig:exp_target_return}, we compare the distribution $ P(\text{expert}_{t} \mid R_{t}, \ldots) $ for sampling target returns during game execution in both Multi-Game DT and \algo, with the distribution of episode returns from offline datasets $ N(R_t) $. 
sample valid target returns that align with the distribution $ N(R_1) $. In other words, Multi-Game DT may encounter issues in action inference because it inputs returns not included in the training data to the Transformer. However, \algo~can promote expert actions by focusing on feasible target returns.


\begin{table}[t]
    \centering
    \caption{Comparison of the number of games with performance improvements and declines across different methods. This table shows the number of games where each method achieves more than 10\% higher or lower IQM compared to the Multi-Game DT.}
    \begin{tabular}{lcc}
        \toprule
        Method & Games Improved & Games Declined \\
        \midrule
        Na\"ive method & 12 & 7 \\
        \algo~(w/o \derd) & 8 & 12 \\
        \algo~(w/o \barp) & 12 & 9 \\
        \algo & 14 & 4 \\
        \bottomrule
    \end{tabular}
    \label{fig:performance_comparison}
\end{table}



\section{Conclusion}
This paper proposes an algorithm called Multi-Game Target Return Optimizer (\algo) to compute optimal target returns for each task within the Multi-Game Decision Transformer framework.  Traditionally, determining target returns required applying human knowledge for each task. By leveraging environmental reward information from offline datasets, \algo~addresses the challenge of identifying suitable target returns necessary for generating expert actions, eliminating the reliance on human expertise. Furthermore, \algo~refines the expert distribution using offline datasets based on the prediction accuracy of the decision transformer. Our approach enhances the performance of pre-trained models through several modifications without requiring additional training. Extensive experiments on Atari games have confirmed that \algo~is effective across a diverse range of games.

\paragraph{Limitations.}
Our approach requires selecting expert data from a large amount of offline datasets to calculate the target return that generates expert actions. However, there may be cases where expert data is absent for some tasks within the offline datasets. Additionally, while this paper considers the experiences from the final stages of a DQN agent's training as expert data, there is room for discussion on the method used to determine how to qualify as expert data.

\paragraph{Future work.}
Our approach is not limited to the multi-game settings and has the potential to enhance the performance of the Decision Transformer framework more broadly. Exploring its application in various offline reinforcement learning algorithms based on Decision Transformers could yield valuable insights and improvements.

In this study, we focused on generating expert actions. However, with minor modifications to \algo, it would be possible to generate medium-level actions as well. Future work could explore these modifications to expand the range of behaviors that can be effectively generated, thereby broadening the applicability and utility of our approach in diverse reinforcement learning scenarios.

\bibliographystyle{named}
\bibliography{ijcai24}
\clearpage
\newpage

\begin{table*}[t]
\caption{Results for the Atari games, including the raw score with mean, standard deviation over 50 trials.}
\begin{center}
\resizebox{\textwidth}{!}{%
\begin{tabular}{lrrrrr}
\toprule
Game Name& Multi-Game DT & Na\"ive method & \algo~(w/o \barp) & \algo~(w/o \derd) & \algo \\
\midrule
Amidar & 144.82 ± 78.09 & 165.24 ± 80.70 & 168.78 ± 87.09 & 172.06 ± 77.03 & 143.04 ± 50.83 \\
Assault & 2106.66 ± 855.90 & 2249.40 ± 908.68 & 2161.32 ± 929.25 & 2302.24 ± 906.58 & 2021.68 ± 918.48 \\
Asterix & 7101.00 ± 1926.80 & 8486.00 ± 2584.72 & 8658.00 ± 4034.60 & 7435.00 ± 2076.30 & 8370.00 ± 3119.89 \\
Atlantis & 76506.00 ± 160720.03 & 264378.00 ± 308614.53 & 264378.00 ± 308614.53 & 8424.00 ± 3253.59 & 251788.00 ± 318011.62 \\
BankHeist & 7.80 ± 9.44 & 6.60 ± 8.86 & 13.80 ± 15.48 & 8.40 ± 13.62 & 7.00 ± 8.54 \\
BattleZone & 14400.00 ± 6764.61 & 17160.00 ± 5927.43 & 18080.00 ± 7282.42 & 15160.00 ± 6640.36 & 17160.00 ± 6197.94 \\
BeamRider & 8003.60 ± 4741.12 & 8022.48 ± 4256.97 & 8034.84 ± 4181.35 & 8116.00 ± 4033.65 & 7828.24 ± 4444.54 \\
Boxing & 91.30 ± 5.80 & 91.24 ± 5.74 & 91.06 ± 6.45 & 92.22 ± 5.13 & 92.20 ± 5.02 \\
Breakout & 246.32 ± 82.49 & 300.16 ± 83.35 & 306.74 ± 96.76 & 243.24 ± 108.50 & 327.30 ± 79.38 \\
Centipede & 2319.92 ± 1662.38 & 2729.88 ± 1569.62 & 2908.16 ± 1915.28 & 2859.08 ± 1843.11 & 2629.30 ± 1466.57 \\
ChopperCommand & 1778.00 ± 913.96 & 1474.00 ± 897.06 & 1484.00 ± 891.37 & 1126.00 ± 755.73 & 1386.00 ± 867.64 \\
CrazyClimber & 121208.00 ± 19079.16 & 126452.00 ± 19331.00 & 126452.00 ± 19331.00 & 128466.00 ± 17821.96 & 126452.00 ± 19331.00 \\
DemonAttack & 12738.60 ± 7340.09 & 10635.60 ± 5667.18 & 10902.20 ± 6300.36 & 11384.50 ± 6706.21 & 11774.30 ± 9930.65 \\
DoubleDunk & -13.80 ± 4.75 & -15.16 ± 5.12 & -13.00 ± 5.05 & -16.34 ± 4.31 & -15.64 ± 5.17 \\
Enduro & 1292.62 ± 276.32 & 1393.82 ± 320.71 & 1393.82 ± 320.71 & 1429.28 ± 269.60 & 1393.82 ± 320.71 \\
FishingDerby & -2.20 ± 34.35 & -6.74 ± 36.97 & -13.50 ± 39.10 & -8.32 ± 38.90 & -11.30 ± 39.16 \\
Freeway & 29.86 ± 1.67 & 29.68 ± 1.55 & 29.74 ± 1.57 & 25.02 ± 1.93 & 29.90 ± 1.53 \\
Frostbite & 2189.20 ± 857.12 & 616.60 ± 617.98 & 528.20 ± 572.80 & 1519.60 ± 848.26 & 876.20 ± 636.09 \\
Gopher & 9869.20 ± 5165.01 & 11724.40 ± 6522.97 & 9370.80 ± 5582.47 & 11698.40 ± 6823.07 & 12218.00 ± 8236.31 \\
Gravitar & 62.00 ± 127.89 & 42.00 ± 109.25 & 45.00 ± 108.28 & 44.00 ± 92.00 & 40.00 ± 104.40 \\
Hero & 19142.80 ± 2538.41 & 19766.60 ± 2333.44 & 20173.40 ± 1322.83 & 20339.90 ± 927.05 & 20319.40 ± 958.41 \\
IceHockey & -9.36 ± 3.33 & -7.88 ± 3.50 & -7.74 ± 3.49 & -10.68 ± 4.00 & -8.56 ± 3.86 \\
Jamesbond & 498.00 ± 268.51 & 683.00 ± 222.40 & 610.00 ± 203.47 & 389.00 ± 287.28 & 641.00 ± 203.15 \\
Kangaroo & 11784.00 ± 2873.84 & 11598.00 ± 3302.03 & 11424.00 ± 3524.23 & 5120.00 ± 3919.29 & 11278.00 ± 2950.21 \\
Krull & 2686.00 ± 782.78 & 1291.00 ± 493.70 & 2649.60 ± 993.43 & 8406.80 ± 1307.64 & 8375.00 ± 1391.07 \\
KungFuMaster & 15136.00 ± 6454.48 & 16134.00 ± 6035.81 & 14144.00 ± 5111.48 & 15152.00 ± 5831.03 & 14690.00 ± 4955.81 \\
NameThisGame & 7705.20 ± 1549.92 & 7722.60 ± 1681.26 & 7722.60 ± 1681.26 & 8230.80 ± 1567.91 & 7722.60 ± 1681.26 \\
Phoenix & 4968.40 ± 486.19 & 5004.20 ± 420.69 & 4926.60 ± 876.75 & 5057.40 ± 490.61 & 4898.60 ± 482.68 \\
Qbert & 6065.00 ± 3179.56 & 5222.50 ± 2655.90 & 4670.00 ± 2951.43 & 6371.50 ± 3043.21 & 6257.00 ± 3341.90 \\
Riverraid & 14690.40 ± 3461.94 & 14236.20 ± 3671.93 & 14648.00 ± 3292.59 & 14973.60 ± 3435.19 & 15432.00 ± 3456.39 \\
RoadRunner & 28264.00 ± 8304.04 & 26334.00 ± 7923.17 & 24340.00 ± 8965.02 & 25240.00 ± 6282.23 & 29580.00 ± 9690.94 \\
Robotank & 49.34 ± 10.03 & 49.28 ± 9.68 & 50.06 ± 8.74 & 44.14 ± 13.90 & 47.74 ± 11.84 \\
Seaquest & 4689.20 ± 1542.62 & 4008.20 ± 2124.38 & 2971.40 ± 1805.58 & 2267.40 ± 1825.26 & 3866.60 ± 1730.78 \\
TimePilot & 2518.00 ± 1713.91 & 3244.00 ± 1637.21 & 2734.00 ± 1624.02 & 2646.00 ± 1645.50 & 2824.00 ± 1634.20 \\
UpNDown & 14261.60 ± 6293.27 & 13609.80 ± 5384.90 & 11841.60 ± 5461.39 & 13435.40 ± 6291.46 & 14550.60 ± 6268.76 \\
VideoPinball & 144.10 ± 399.25 & 278.40 ± 551.28 & 208.38 ± 479.88 & 378.40 ± 803.62 & 318.34 ± 718.69 \\
WizardOfWor & 358.00 ± 380.57 & 418.00 ± 418.42 & 354.00 ± 280.86 & 380.00 ± 364.97 & 388.00 ± 340.38 \\
YarsRevenge & 15625.96 ± 8724.59 & 15733.74 ± 8848.26 & 14683.52 ± 8988.26 & 15266.96 ± 7967.17 & 17672.94 ± 10656.70 \\
Zaxxon & 232.00 ± 422.58 & 164.00 ± 243.93 & 76.00 ± 142.21 & 60.00 ± 154.92 & 306.00 ± 349.52 \\

\bottomrule
\end{tabular}
}
\label{tab:raw_score}
\end{center}
\end{table*}
\end{document}